\documentclass[10pt,twocolumn,letterpaper]{article}

\usepackage{cvpr}              
\usepackage[table]{xcolor}
\usepackage{url}
\usepackage{etoc}
\usepackage[accsupp]{axessibility} 
\usepackage{algorithm}       
\usepackage{algpseudocode}
\usepackage{booktabs}   
\usepackage{graphicx}   
\usepackage{multirow}
\usepackage{pifont}
\usepackage{subcaption}
\usepackage{wrapfig}
\usepackage{amsthm}
\usepackage{enumitem}

\usepackage{marvosym}

\newcommand{\cmark}{\ding{51}} 
\newcommand{\xmark}{\ding{55}} 

\definecolor{GreenCheck}{RGB}{0, 102, 51}

\definecolor{customblue}{HTML}{2C70C9}
\definecolor{customred}{HTML}{E74A7A}
\definecolor{imgtokenblue}{HTML}{14B6F1}

\definecolor{rebeccapurple}{HTML}{663399}

\definecolor{cvprblue}{rgb}{0.21,0.49,0.74}
\usepackage[pagebackref,breaklinks,colorlinks,allcolors=cvprblue]{hyperref}
\usepackage{booktabs,multirow}
\def\confName{CVPR}
\def\confYear{2026}

\title{E3AD: An Emotion-Aware Vision-Language-Action Model for Human-Centric End-to-End Autonomous Driving}

\author{
    \textbf{Yihong Tang}$^{1}$\textsuperscript{\href{mailto:yihong.tang@mail.mcgill.ca}{\Cancer}},~
    \textbf{Haicheng Liao}$^{2}$\textsuperscript{\href{mailto:yc27979@um.edu.mo}{\Cancer}},~
    \textbf{Tong Nie}$^{3}$,~
    \textbf{Junlin He}$^{3}$,~
    \textbf{Ao Qu}$^{4}$,~
    \textbf{Kehua Chen}$^{5}$,~
    \textbf{Wei Ma}$^{3}$,~ \\
    \textbf{Zhenning Li}$^{2}$\textsuperscript{\href{mailto:zhenningli@um.edu.mo}{\Letter}},~
    \textbf{Lijun Sun}$^{1}$\textsuperscript{\href{mailto:lijun.sun@mcgill.ca}{\Letter}},~
    \textbf{Chengzhong Xu}$^{2}$\textsuperscript{\href{mailto:czxu@um.edu.mo}{\Letter}}  \\
    $^1$McGill University \quad
    $^2$University of Macau \quad
    $^3$The Hong Kong Polytechnic University \\ 
    $^4$Massachusetts Institute of Technology \quad
    $^5$University of Washington \\
    \vspace{-2mm}
    \small{
    \href{mailto:yihong.tang@mail.mcgill.ca}{\texttt{yihong.tang@mail.mcgill.ca}} \quad
    \href{mailto:czxu@um.edu.mo}{\texttt{\{yc27979,zhenningli,czxu\}@um.edu.mo}} \quad
    \href{mailto:lijun.sun@mcgill.ca}{\texttt{lijun.sun@mcgill.ca}}
    }
}

\begin{document}
\maketitle
\begingroup
\renewcommand\thefootnote{}
\footnotetext{\textsuperscript{\Cancer} Equal contribution. \, \textsuperscript{\Letter} Corresponding Authors.}
\endgroup

\begin{abstract}
End-to-end autonomous driving (AD) systems increasingly adopt vision-language-action (VLA) models, yet they ignore the passenger’s emotional state, which is central to comfort and AD acceptance. We introduce Open-Domain End-to-End (OD-E2E) AD, where an autonomous vehicle must interpret free-form natural-language commands, infer the emotion, and plan a physically feasible trajectory. We propose E3AD, an emotion-aware VLA framework that augments semantic understanding with two cognitively inspired components: a continuous Valence-Arousal-Dominance (VAD) emotion model that captures tone and urgency from language, and a dual-pathway spatial reasoning module that fuses egocentric and allocentric views for human-like spatial cognition. A consistency-oriented training scheme, combining modality pretraining with preference-based alignment, further enforces coherence between emotional intent and driving actions. Across real-world datasets, E3AD improves visual grounding and waypoint planning and achieves state-of-the-art (SOTA) VAD correlation for emotion estimation. These results show that injecting emotion into VLA-style driving yields more human-aligned grounding, planning, and feedback.
\end{abstract}

\section{Introduction}\label{sec:intro}
\vspace{-2mm}

Autonomous driving (AD) has evolved from modular pipelines to vision-language-action end-to-end (E2E) systems that directly map sensor inputs to vehicle controls through unified optimization. This paradigm significantly improves efficiency and adaptability by integrating perception, prediction, and planning into a single learning framework~\cite{li2024steering,chen2024end}. Despite these advances, a fundamental obstacle remains that is not purely technical but human-centered: ensuring public trust and acceptance of fully AD~\cite{zhao2025survey}.

\begin{figure}[t]
  \centering
    \vspace{-2.5mm}
    \includegraphics[width=.95\linewidth]{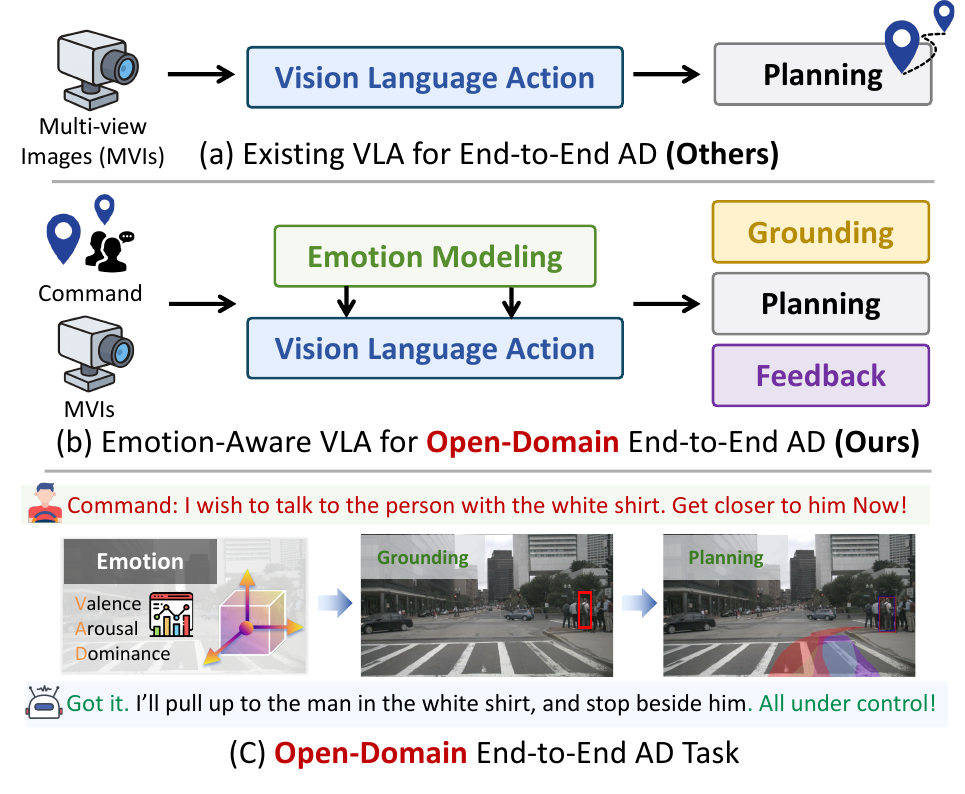}
    \vspace{-4mm}
    \caption{Overview of our E3AD. (a) Existing VLAs behave as emotion-agnostic systems, mapping multi-view images directly to a planning output without human-in-the-loop interaction or emotion understanding. (b) Our model adds explicit emotion modeling and closed-loop feedback, allowing the agent to infer intent intensity, ground referents more reliably, and adapt its plan accordingly. (c) This yields the Open-Domain E2E AD task, where the agent jointly reasons over language, emotion, perception, and navigation to enable human-centered and context-aware autonomy.}
    \label{head}
    \vspace{-6mm}
\end{figure}

While current E2E systems~\cite{hu2023planning,prakash2021multi,jia2023think} exhibit strong control and perception capabilities, passengers often feel uneasy about delegating decisions to opaque algorithms that operate without acknowledging human intent or emotion. Surveys~\cite{zepf2020driver} and behavioral studies~\cite{chen2025emotion,liao2025toward,yang2025drivemoe} consistently indicate that emotional interaction is a critical determinant of user comfort and perceived safety. However, most existing models~\cite{li2025drivevla,nie2025steerable} are designed for closed-loop rational control and remain insensitive to emotion cues such as anxiety or urgency. This disconnection between computational reasoning and emotional understanding forms what can be described as an \textit{emotion gap} for autonomous vehicles (AVs). Bridging this gap requires reconsidering the role of human-vehicle interaction in AD. An intelligent system should understand ``\textit{what}'' a passenger says and ``\textit{how}'' it is expressed. For instance, the tone difference between ``stop here'' and ``stop here now!'' carries implicit emotional meaning that influences how the vehicle should respond. Recognizing such differences enables the system to regulate behavior that aligns with the passenger’s emotional state, providing reassurance and enhancing acceptance~\cite{jiang2025survey}.

As illustrated in Fig.~\ref{head}, we extend the conventional E2E AD framework toward a more human-centric paradigm. Future AVs must reason not only over visual and spatial cues but also interpret and respond to natural-language commands that convey the passenger’s intent and emotional state. We define this capability as the task of \textbf{O}pen-\textbf{D}omain End-to-End AD (\textbf{OD-E2E}), where the driving agent jointly reasons over semantic content, emotion context, and spatial environment to generate physically realizable trajectories consistent with the passenger’s instructions. This formulation moves beyond purely reactive control toward interactive, emotion-aware driving assistants that better match human expectations and preferences, transforming AVs into empathetic and user-aligned driving agents~\cite{li2024steering}.

Technically, our approach builds on the emerging Vision-Language-Action (VLA) paradigm~\cite{sapkota2025vision}, which unifies perception, reasoning, and control through large-scale multimodal modeling. By coupling visual and linguistic representations, VLA frameworks enable agents to perform complex goal-driven behaviors with improved generalization, interpretability, and alignment to human intent. Following this paradigm, we propose an \textbf{E}motion-aware \textbf{E}nd-to-\textbf{E}nd \textbf{A}utonomous \textbf{D}riving framework, \textbf{E3AD}, which extends VLA from purely semantic understanding to emotion and spatially consistent reasoning. To enhance cognitive capability, E3AD incorporates two key components within a unified pipeline: (1) Emotion Modeling, which maps commands into a continuous Valence-Arousal-Dominance (VAD) space~\cite{gaertner2021multi} to interpret emotional tone and behavioral urgency; and (2) Spatial Reasoning, which fuses egocentric and allocentric pathways to achieve human-like spatial cognition.  These components are jointly optimized through a consistency-oriented learning strategy that enforces coherence between the semantic and emotional context of the command and the resulting trajectory. This design enables E3AD to reason jointly over \textit{what} the passenger intends and \textit{how} it is expressed, producing emotion-aware and human-aligned driving behaviors in a fully E2E manner. From a safety perspective, modulating behavior according to passenger state decouples request satisfaction from the safety envelope, which is essential for calibrated trust in E2E AD.

Overall, the contributions of this study are threefold:
\begin{itemize}[leftmargin=*]
    \item We define Open-Domain End-to-End AD for human-centric AVs, which unifies semantic, emotional, and spatial reasoning from natural-language commands.
    
    \item  We propose E3AD, an emotion-aware VLA framework that integrates continuous emotion modeling and dual-system spatial reasoning in a unified E2E pipeline, enabling emotion-grounded response and planning.
    
    \item  We show that E3AD outperforms strong baselines on visual grounding, emotion estimation, and waypoint planning across multiple benchmarks, with particularly large gains on emotion-sensitive and corner-case scenarios.
\end{itemize}

\vspace{-1mm}
\section{Related Work}
\vspace{-0.3mm}

\noindent\textbf{VLA for End-to-end AD.}
Recent work explores VLA architectures that inject world knowledge and reasoning from large multimodal models into E2E AD~\cite{cui2024survey}. Existing approaches can be broadly grouped into three paradigms~\cite{jiang2025survey}. The first paradigm, represented by DriveGPT-4~\cite{xu2024drivegpt4}, OpenEMMA~\cite{xing2025openemma}, and CoT-Drive~\cite{liao2025cot}, produces scene-level explanations via QA-style prompts. While improving interpretability, they act as ``commentators'', lacking precise spatial grounding and direct control fidelity. The second paradigm, such as Senna~\cite{jiang2024senna}, VLP~\cite{pan2024vlp}, and LMDrive \cite{shao2024lmdrive}, employs VLMs to generate discrete ``meta-behaviors'' to guide a low-level controller. This approach provides only sparse guidance that limits its capacity for continuous spatial reasoning, resulting in marginal gains in driving performance. The third paradigm, like Simlingo~\cite{renz2025simlingo}, AutoVLA~\cite{zhou2025autovla}, and FSDrive~\cite{zeng2025futuresightdrive}, couples VLM-based perception with dedicated planning modules that directly output trajectories or control signals, achieving strong performance. However, current VLAs~\cite{liu2025omnireason,li2025fine,zhou2025opendrivevla} face two core issues: weak spatial understanding, operating largely in 2D without explicit 3D or allocentric (map-based) reasoning~\cite{tang2024itinera,tang2025sparkle}, and a purely rational sequence-prediction view that ignores passenger emotion, crucial for behavior alignment~\cite{tang2026intent,tang2025routekg}. Building on the third paradigm, we propose an emotion-aware, spatially grounded end-to-end VLA.

\begin{figure*}[htbp]
\centering
\includegraphics[width=1\textwidth]{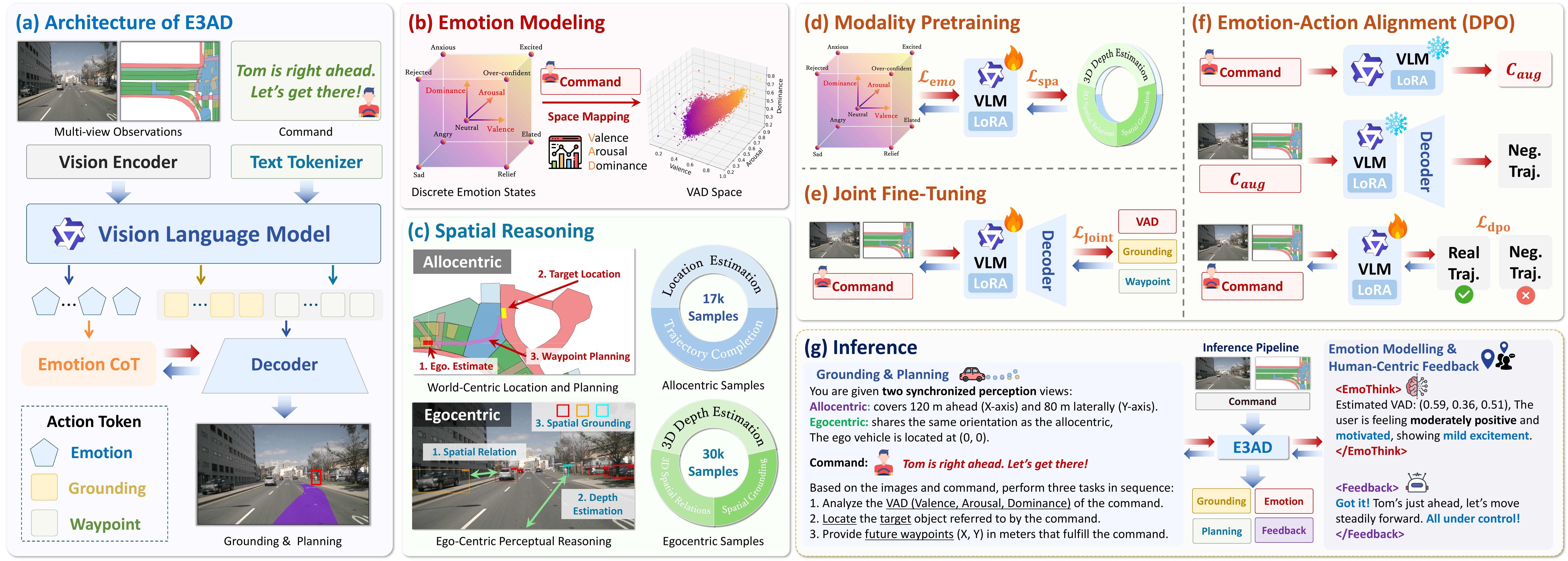}
\captionsetup{skip=2pt}
\caption{Overview of E3AD and its training/inference pipeline. Given egocentric and allocentric views with a natural-language command (a), E3AD outputs emotion, grounding, and waypoint tokens via two core modules: Emotion Modeling (b) encodes commands in continuous VAD space (c), and Spatial Reasoning fuses egocentric and allocentric pathway cues. Training proceeds from Modality Pretraining for emotion/spatial skills (d) to Joint Fine-Tuning that predicts ($\hat{e}$, $\hat{b}$, $\hat{\tau}$) in a single autoregressive chain (e), followed by Emotion-Action Alignment (f). During inference (g), E3AD runs end-to-end to estimate ($\hat{e}$), ground ($\hat{b}$), and plan ($\hat{\tau}$), producing human-centric feedback.}
\vspace{-2mm}
\label{fig:e3ad}
\end{figure*}

\noindent\textbf{Emotion Computing in AD.}
Emotional interaction between humans and AVs offers a promising path to improving public acceptance, safety, and comfort in AD~\cite{li2024review}. However, enabling AVs to accurately perceive, interpret, and respond to human emotions remains a fundamental challenge~\cite{sini2020passengers}. Early work~\cite{cowie2001emotion,mou2023driver} emphasized driver-state monitoring (fatigue, distraction, stress) via physiological signals (EEG/ECG~\cite{chen2024eeg}) and visual cues (facial expressions~\cite{xiao2022road} or gaze tracking); with higher-level AD, the focus shifted to passenger experience (e.g., ride comfort \cite{liao2024gpt,liang2025mamba}, anxiety \cite{lee2020detection,xiang2025driver}). Most approaches perform passive detection with discrete labels and decouple emotion estimation from downstream control. We instead ground emotion in the continuous VAD space, integrating it into a unified VLA. To our knowledge, E3AD is the first framework to use a VAD vector to guide: (1) ambiguity resolution for nuanced commands, and (2) the generation of planning trajectories. Through consistency-oriented fine-tuning, E3AD aligns emotion with driving behavior, advancing from emotion recognition to a human-centric AD system~\cite{yang2024human}.

\section{Methodology}
\subsection{Problem Formulation}
We consider OD-E2E AD, where an autonomous vehicle receives multi-view observations and a natural-language command and must execute the command through spatial grounding and motion planning. Let \(C\) denote the passenger’s command and \(I=\{I_{\text{ego}}, I_{\text{allo}}\}\) denote the multi-view observations containing egocentric and allocentric views of the scene. The agent must (i) localize the object or location referenced in \(C\) and (ii) generate a physically feasible trajectory that fulfills the instructed intent. Formally, given the tuple $(I, C)$, the task aims to learn a mapping:
\vspace{-2mm}
{\begin{small}
\begin{equation}
    f_\theta: (I, C) \rightarrow 
    \hat{\mathcal{Y}} = \{\hat{b}, \hat{\tau}\},
\end{equation}
\end{small}}
\vspace{-6mm}

\noindent where \(\hat{b}\) denotes the grounded target in the scene and \(\hat{\tau} = \{y_t\}_{t=1}^{T}\) denotes the future waypoints.
Unlike prior language grounding setups that restrict command vocabularies and decouple referent localization from motion planning into separate modules,
the OD-E2E task learns a single policy \(f_\theta\) that, under a unified objective, jointly predicts 
\((\hat{b}, \hat{\tau})\) conditioned on $(I, C)$.
This elevates language grounding from an auxiliary perception task to an integral component of the end-to-end decision-making objective.

\subsection{Overview}
\label{ssec:overview}
Fig.~\ref{fig:e3ad} illustrates the overall architecture of E3AD, a cognitively inspired VLA system built on Qwen2.5-VL-7B-Instruct~\cite{bai2025qwen2} that integrates Emotion Modeling, Two-System Spatial Representation, and Consistency-Oriented Action Planning in a unified pipeline. To strengthen its cognitive competence, we adopt a three-stage training strategy: (i) Foundational Modality Pretraining to establish emotion and spatial priors, (ii) Joint Reasoning Fine-tuning to optimize end-to-end multimodal operation, and (iii) Emotion-Action Alignment to ensure planning behaviors remain consistent with emotional intent. This progressive design provides fine-grained emotion understanding, cognitively grounded spatial reasoning, and behaviorally consistent planning, all aligned with the passenger’s semantic and emotional intent.

\begin{figure}[t]
  \centering
    \includegraphics[width=.8\linewidth]{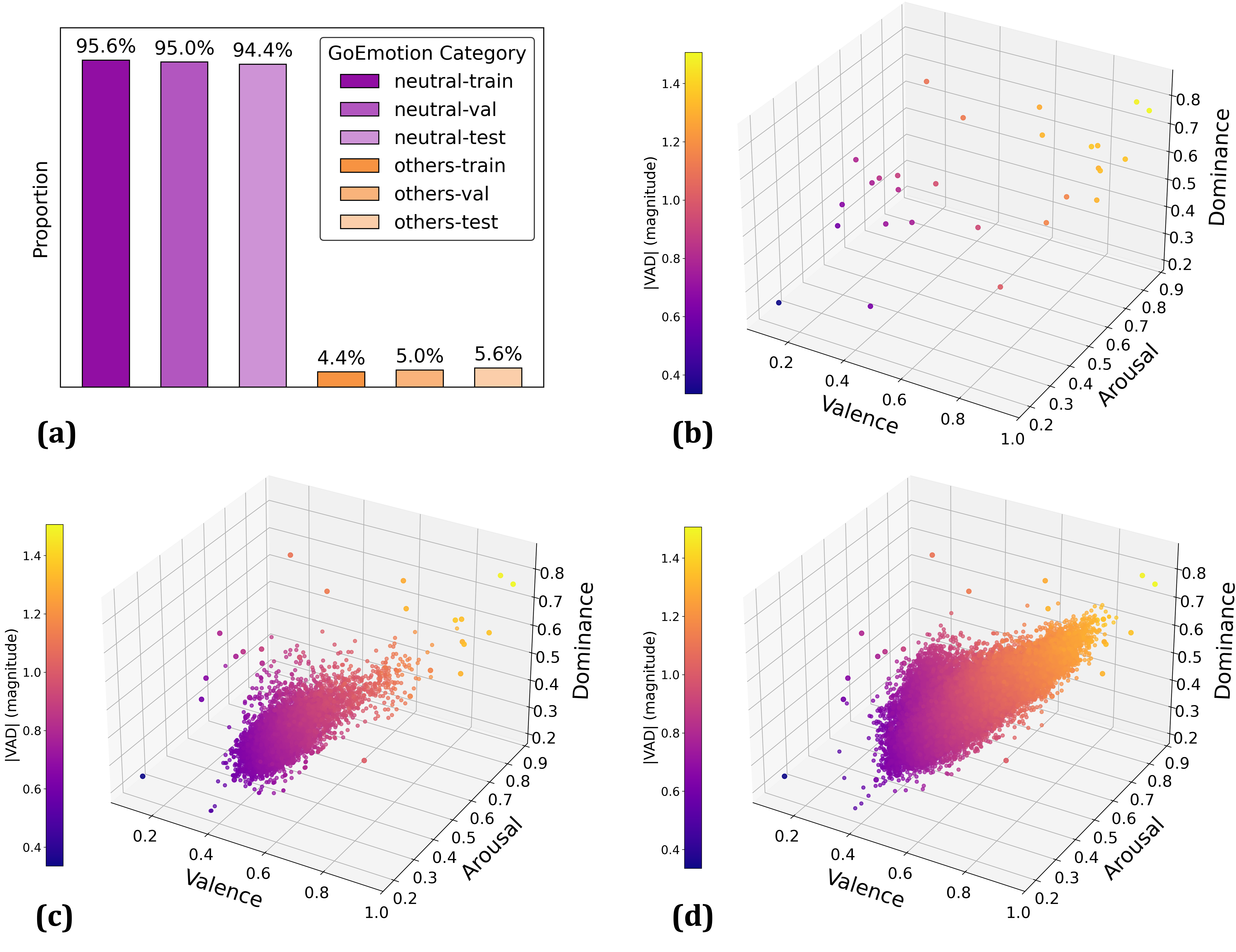}
\vspace{-2mm}
    \caption{Visualization of emotion distributions before and after augmentation. (a) Proportions of GoEmotion categories across Talk2Car splits.
(b) VAD distribution of GoEmotion.
(c) Incorporating driving commands enriches emotional diversity. 
(d) Emotion-aware augmentation expands and smooths the VAD distribution, providing broader and continuous emotion supervision.}
    \label{fig:emotion_aug}
    \vspace{-4mm}
\end{figure}

\subsection{Emotion Modeling}
\label{ssec:emo}

Human-vehicle interaction in AD inherently involves emotional communication. Yet, most existing systems are emotion-agnostic and model affect, if at all, with a small set of discrete labels such as happy, angry, or sad~\cite{liao2024gpt}. Such coarse encoding cannot capture small but behaviorally meaningful tone changes that influence how a command should be executed. We therefore adopt a continuous representation based on the Valence-Arousal-Dominance (VAD) model~\cite{warriner2013norms}, representing emotion as $e \in \mathbb{R}^3$. Valence measures positivity, arousal measures activation, and dominance measures control; in driving contexts, these axes correspond to attitude (calm vs. anxious), alertness (fatigued vs. vigilant), and control (confident vs. overwhelmed)~\cite{matthews1991personality}.

To supervise this space, we derive VAD labels from both sentence-level and word-level cues. For each command $C$, we first apply the GoEmotions classifier~\cite{demszky2020goemotions} to obtain a distribution over discrete emotions and map it to sentence-level VAD scores using the label-VAD dictionary in~\cite{warriner2013norms}. In parallel, we compute a word-level VAD vector by removing stop words and averaging lexical scores over the remaining tokens. The final label $e$ is obtained by combining these two sources, so that both global interpretation and emotion-bearing phrases are reflected (details in Appendix~\textcolor{cvprblue}{A.1}).

Driving commands, however, are often emotionally neutral, which would encourage the model to ignore emotion if trained naively. To break this bias and disentangle intent from tone, we introduce emotion-aware command augmentation. For each command $C^{(i)}$, Qwen2.5-VL generates $K$ paraphrases $C^{(i)}_{\text{aug}} = \{C^{(i)}_1, \ldots, C^{(i)}_K\}$ that preserve the driving goal while varying the attitude or intensity. Each augmented command $C^{(i)}_k$ is assigned a VAD label $e^{(i)}_k$ via the same mapping procedure, forming an augmented set $\mathcal{C}^*$ that creates neighborhoods of semantically equivalent but affectively distinct commands (Fig.~\ref{fig:emotion_aug}). This forces the model to attribute changes in $e$ to changes in tone rather than intent.

We then perform supervised fine-tuning (SFT) to equip the VLM with explicit emotion understanding. Given augmented command-emotion pairs $\{(C^{(i)}_k, e^{(i)}_k)\} \subset \mathcal{C}^*$, we cast emotion prediction as conditional generation with an instruction template (Appendix~\textcolor{cvprblue}{D}) and define the loss as:
\begin{equation}
\mathcal{L}_{\text{emo}}
= -\mathbb{E}_{(C^{(i)}_k, e^{(i)}_k)\sim\mathcal{C}^*}
\big[\log p_{\theta}(e^{(i)}_k \mid C^{(i)}_k)\big],
\end{equation}
where $p_{\theta}$ denotes the model’s distribution over quantized VAD tokens. Rather than adding a separate sentiment head, this continuous, instruction-style formulation embeds $e$ within the same generative reasoning process as other outputs, enabling E3AD to represent fine-grained shifts in affect while keeping the underlying intent fixed, and to condition planning behavior directly on the inferred emotion.

\subsection{Spatial Reasoning}

Inspired by the dual-system model of human spatial perception~\citep{burgess2006spatial}, we design the VLA backbone to reason over two complementary spatial pathways: an \textit{egocentric} frame for immediate, action-oriented perception and an \textit{allocentric} frame for global, map-based structure. This design compels the model to fuse local sensory cues with world-centered priors, mirroring how humans combine first-person observations with internal cognitive maps for reliable navigation.

\noindent\textbf{Egocentric Pathway.}
The egocentric pathway captures the agent’s first-person perceptual field. Using annotated data samples, the model predicts (i) the relative 3D direction to the referent, (ii) its distance, and (iii) its grounded location in image coordinates from $(I_{\text{ego}}, C)$. These auxiliary signals provide fine-grained, short-horizon spatial cues essential for immediate control. We curate a dataset of 30K samples to establish a robust foundation for egocentric spatial reasoning, which directly supports subsequent planning.

\noindent\textbf{Allocentric Pathway.}
The allocentric pathway encodes a world-centered representation akin to a cognitive map. Given BEV input $I_{\text{allo}}$, the model learns to (i) predict the target location in BEV coordinates and (ii) generate a coarse trajectory $\tau=\{y_t\}_{t=1}^{T}$ from the ego pose to that target. This supervision yields long-horizon spatial structure, road topology, occlusions, and multi-agent layout, producing map-consistent priors that complement egocentric perception. We use roughly 17K samples to teach this world-centered reasoning.  In combination, the two pathways supply complementary local and global spatial cues that downstream planning modules can exploit for grounding and waypoint generation in cluttered, partially observed scenes.

\subsection{Action and Feedback}

\noindent\textbf{Action Decoder.}
Following the VLA backbone, we append a lightweight action decoder $f_{\text{act}}$ for translating the VLA's high-level outputs into a precise, physically-realizable trajectory $\hat{\tau}$. Conditioned on the grounded target $\hat{b}$, the coarse trajectory $\widetilde{\tau}$, and visual observations $I$, the decoder outputs the final trajectory $\hat{\tau} = f_{\text{act}}(\hat{b}, \widetilde{\tau}, I)$, where $\hat{\tau} \in \mathbb{R}^{T \times 2}$ represents the spatial coordinates of waypoints.

\noindent\textbf{Human-centric Verbal Feedback.}
To mitigate the ``black-box'' anxiety of passengers, E3AD provides verbal feedback $\hat{r}$ after planning the waypoints. We employ the trained Qwen2.5-VL backbone to generate this response. Specifically, the model is guided by structured prompts, conditioning the generation of $\hat{r}$ on the complete output of the VLA pipeline: the predicted emotion state $\hat{e}$, the grounded target $\hat{b}$, and the planned waypoints $\hat{\tau}$.
The response policy adapts tone, length, and specificity to emotion and urgency: for calm states, it offers brief confirmations, whereas for high arousal, it produces direct, time-critical guidance. This emotion-aware feedback loop transforms the AV from an opaque tool into a human-centric agent.

\subsection{Consistency-Oriented Training and Inference}

We adopt a three-stage consistency-driven training strategy to progressively endow E3AD with emotion awareness, spatial reasoning, and coherent decision-making. The training stages consist of: (1) Modality Pretraining, which establishes foundational representations for spatial and emotional cues; (2) Joint Fine-tuning, unifying emotion grounding, scene understanding, and trajectory planning within a single autoregressive generation process; and (3) Emotion-Action Alignment, designed for stable, emotion-consistent driving behaviors. Each training stage is as follows:

\renewcommand{\arraystretch}{1.1}
\begin{table*}[t]
\centering
\caption{End-to-end performance of E3AD vs. state-of-the-art (SOTA) baselines. Best results are \textbf{bold}; second-best are \underline{underlined}.}
\vspace{-2.5mm}
\setlength{\tabcolsep}{10pt}
\resizebox{.95\linewidth}{!}{
\begin{tabular}{l|ccccc|cc}
\bottomrule
\textbf{Model} & \textbf{ADE $\downarrow$} & \textbf{Fréchet $\downarrow$} & \textbf{SSPD $\downarrow$} & \textbf{DTW $\downarrow$} & \textbf{FDE $\downarrow$} & $\mathbf{PA_2}$ $\uparrow$ & $\mathbf{PA_4}$ $\uparrow$ \\
\hline
$A^*$-ROL~\cite{hart1968formal} & $5.63 \pm 0.12$ & $10.22 \pm 0.20$ & $3.06 \pm 0.06$ & $93.35 \pm 1.80$ & $9.34 \pm 0.18$ & $5.45 \pm 1.38$ & $25.13 \pm 2.63$ \\
$A^*$-PE (PDPC)~\cite{hart1968formal} & $5.35 \pm 0.13$ & $9.38 \pm 0.22$ & $2.80 \pm 0.07$ & $86.78 \pm 2.28$ & $8.22 \pm 0.21$ & $25.95 \pm 2.66$ & $46.41 \pm 3.03$ \\
GoalGAN~\cite{dendorfer2020goal} & $5.89 \pm 0.12$ & $10.86 \pm 0.21$ & $3.09 \pm 0.06$ & $98.75 \pm 2.11$ & $10.08 \pm 0.20$ & $17.32 \pm 2.30$ & $33.82 \pm 2.87$ \\
PECNet~\cite{zhao2021you} & $4.78 \pm 0.10$ & $8.84 \pm 0.19$ & $2.54 \pm 0.05$ & $76.87 \pm 1.75$ & $8.22 \pm 0.17$ & $15.59 \pm 2.20$ & $37.68 \pm 2.94$ \\
Y-net~\cite{mangalam2021goals} & $5.28 \pm 0.10$ & $10.01 \pm 0.19$ & $2.50 \pm 0.04$ & $85.25 \pm 1.69$ & $8.98 \pm 0.18$ & $2.49 \pm 0.94$ & $30.84 \pm 2.81$ \\
TTST~\cite{mangalam2021goals} & $5.24 \pm 0.10$ & $9.79 \pm 0.19$ & $2.47 \pm 0.04$ & $84.02 \pm 1.68$ & $8.50 \pm 0.18$ & $13.98 \pm 2.11$ & $37.54 \pm 2.94$ \\
CWS~\cite{mangalam2021goals} & $4.82 \pm 0.11$ & $9.30 \pm 0.20$ & $2.46 \pm 0.05$ & $78.69 \pm 1.84$ & $8.59 \pm 0.18$ & $3.27 \pm 1.08$ & $34.66 \pm 2.89$ \\
TTST + CWS~\cite{mangalam2021goals} & $4.76 \pm 0.10$ & $8.95 \pm 0.19$ & $2.41 \pm 0.05$ & $76.84 \pm 1.77$ & $8.08 \pm 0.17$ & $17.54 \pm 2.31$ & $40.79 \pm 2.96$ \\
PTPC~\cite{grujicic2022predicting} & \color{brown}\underline{$4.54 \pm 0.11$} & \color{brown}\underline{$8.55 \pm 0.20$} & {\color{brown}\underline{$2.18 \pm 0.05$}} & {\color{brown}\underline{$72.09 \pm 1.86$}} & {\color{brown}\underline{$7.75 \pm 0.19$}} & {\color{brown}\underline{$24.46 \pm 2.61$}} & {\color{brown}\underline{$45.55 \pm 3.03$}} \\

\hline
Qwen2.5-VL-72B~\cite{bai2025qwen2} & $12.51 \pm 0.28$ & $26.15 \pm 0.52$ & $5.87 \pm 0.14$ & $206.99 \pm 4.80$ & $25.85 \pm 0.55$ & $1.18 \pm 0.42$ & $3.13 \pm 0.60$ \\
Qwen3-VL-8B~\cite{yang2025qwen3} & $14.07 \pm 0.30$ & $28.38 \pm 0.58$  & $6.65 \pm 0.16$  & $234.81 \pm 5.20$  & $27.96 \pm 0.60$  & $1.89 \pm 0.50$  & $4.50 \pm 0.70$ \\
FSDrive-Finetuned~\cite{zeng2025futuresightdrive} & $5.02 \pm 0.12$ & $10.98 \pm 0.24$ & $2.28 \pm 0.06$ & $74.50 \pm 2.10$ & $10.45 \pm 0.22$ & $17.10 \pm 2.40$ & $27.85 \pm 2.80$ \\

CAVG (+Planner)~\cite{liao2024gpt} & $4.88 \pm 0.14$ & $9.23 \pm 0.20$ & $2.20 \pm 0.04$ & $73.51 \pm 1.42$ & $9.23 \pm 0.28$ & $20.10 \pm 2.45$ & $43.25 \pm 2.63$ \\ 
\hline 

\multirow{2}{*}{\textbf{E3AD (Ours)}} & \cellcolor{gray!10}\textbf{3.88 $\pm$ 0.10} & 
\cellcolor{gray!10}\textbf{7.23 $\pm$ 0.19} & 
\cellcolor{gray!10}\textbf{1.86 $\pm$ 0.05} & 
\cellcolor{gray!10}\textbf{60.07 $\pm$ 1.69} & 
\cellcolor{gray!10}\textbf{6.64 $\pm$ 0.18} & 
\cellcolor{gray!10}\textbf{36.21 $\pm$ 0.94} & 
\cellcolor{gray!10}\textbf{55.62 $\pm$ 0.95} \\

 & 
\cellcolor{gray!25}$\textbf{17.01\%}\uparrow$ &
\cellcolor{gray!25}$\textbf{18.26\%}\uparrow$ &
\cellcolor{gray!25}$\textbf{17.20\%}\uparrow$ &
\cellcolor{gray!25}$\textbf{16.67\%}\uparrow$ &
\cellcolor{gray!25}$\textbf{20.00\%}\uparrow$ &
\cellcolor{gray!25}$\textbf{16.71\%}\uparrow$ &
\cellcolor{gray!25}$\textbf{18.10\%}\uparrow$ \\
\toprule
\end{tabular}
}
\label{tab:baselines}
\vspace{-2mm}
\end{table*}

\noindent\textbf{Stage-1: Modality Pretraining.} 
In this initial stage, we apply supervised fine-tuning to equip E3AD with spatial and emotion perception skills separately: (1) Emotion Modeling is trained on our augmented command dataset $\mathcal{C}^*$ using the emotion regression loss $\mathcal{L}_\mathrm{emo}$. (2) Spatial Reasoning is trained on our synthetic egocentric and allocentric pathways using the negative log-likelihood loss $\mathcal{L}_\mathrm{spatial}$ of next-token prediction to optimize spatial reasoning across both views.

\noindent\textbf{Stage-2: Joint Fine-Tuning.}
After modality pretraining, we fine-tune the VLA to unify these capabilities into a single, coherent reasoning process. We employ an instruction-based SFT paradigm where the model autoregressively predicts the full output sequence $\mathcal{T} = (\hat{e}, \hat{b}, \hat{\tau})$ in a single forward pass. This joint objective $\mathcal{L}_{\text{joint}}$ is defined as follows:

\vspace{-3mm}
{\begin{small}
\begin{equation}
\mathcal{L}_{\text{Joint}}
= -\mathbb{E}_{(I, C, T)}
\sum_{t=1}^{|T|} \log p_\theta(T_t \mid T_{<t}, I, C),
\end{equation}
\end{small}}

\vspace{-1mm}
\noindent where $p_\theta$ is the conditional token distribution. This encourages the VLA to form an emotion-aware chain of thought, where the predicted emotion $\hat{e}$ and spatial grounding $\hat{b}$ directly inform the subsequent generation of waypoints $\hat{\tau}$.

\noindent\textbf{Stage-3: Emotion-Action Alignment (DPO).}
While $\mathcal{L}_{\text{Joint}}$ aligns tasks, it does not explicitly enforce behavioral consistency with different emotional intent. Standard preference-based alignment methods (like DPO) are non-trivial to apply in this field, as AD datasets typically provide only a single, actual trajectory $\tau^{(i)}$ for any given command $C^{(i)}$, rather than ranked preference pairs. To address this, we construct a dataset of \textit{pseudo-preference pairs} via emotion-augmented commands. For each original command $C^{(i)}$ and its actual trajectory $\tau^{(i)}$, we identify an emotion-augmented variant $C^{(i)}_{k^-}$ whose VAD embedding deviates most from the original. We use this ``negative'' command to generate a dispreferred, emotion-shifted trajectory $\widetilde{\tau}^{(i)}_{k^-}$:

\vspace{-4mm}
{\begin{small}
\begin{equation}
\label{eq:negativesample}
\begin{aligned}
C^{(i)}_{k^-} = \arg\max_k \| e^{(i)}_k - e^{(i)} \|_2, 
\widetilde{\tau}^{(i)}_{k^-} \sim p_\theta(\tau \mid C^{(i)}_{k^-}, I^{(i)}),
\end{aligned}
\end{equation}
\end{small}
}
\vspace{-4.5mm}

\noindent This yields a preference pair $(\tau^{(i)} \succ \widetilde{\tau}^{(i)}_{k^-})$ for the command $C^{(i)}$. We apply DPO~\cite{rafailov2023direct} to optimize this preference:

\vspace{-1mm}
{\begin{small}
\begin{equation}
\label{eq:dpo}
\begin{aligned}
\mathcal{L}_{\text{dpo}}
= -\mathbb{E}_{i}\Big[
\log \sigma\!\Big(\beta \Big(
&\log p_\theta(\tau^{(i)} \mid C^{(i)}) \\
- &\log p_\theta(\widetilde{\tau}^{(i)}_{k^-} \mid C^{(i)})
\Big)\Big)\Big].
\end{aligned}
\end{equation}
\end{small}}

\vspace{-0.5mm}
\noindent Notably, this stage encourages the model to assign higher likelihood to trajectories consistent with the original command's intent while suppressing emotionally perturbed alternatives, leading to stable yet emotion-aware behavior.

\noindent\textbf{Inference.} After the VLA backbone is trained, its outputs are integrated into a lightweight action decoder for precise trajectory waypoint generation.
During inference, E3AD operates end-to-end: it seamlessly processes the input tuple $(I_\text{ego}, I_\text{allo}, C)$ to directly produce the emotion state $\hat{e}$, the grounded target $\hat{b}$, coarse trajectory waypoints $\hat{\tau}$, and the verbal response $\hat{r}$ for passengers. This integrated process enables emotion-grounded visual grounding and human-aligned planning without additional post-processing.

\renewcommand{\arraystretch}{1.1}
\begin{table*}[t]
\centering
\vspace{-1mm}
\caption{Comparison of E3AD and state-of-the-art baselines on visual grounding tasks. Best results are \textbf{bold}; second-best are \underline{underlined}.}
\vspace{-2.5mm}
\setlength{\tabcolsep}{5pt}
\resizebox{.95\linewidth}{!}{
\begin{tabular}{l| l| c| cc| cc| ccc| c}
\bottomrule
\multirow{2}{*}{\textbf{Model}} & \multirow{2}{*}{\textbf{Backbone}} & \multirow{2}{*}{\textbf{Talk2Car}} &
\multicolumn{2}{c|}{\textbf{MoCAD}} &
\multicolumn{2}{c|}{\textbf{DrivePilot}} &
\multicolumn{3}{c|}{\textbf{Corner-case Test sets}} &
\multicolumn{1}{c}{\textbf{Long-text}} \\
\cmidrule(lr){4-5}\cmidrule(lr){6-7}\cmidrule(lr){8-10}\cmidrule(lr){11-11}
& & & \textbf{test} & \textbf{val} & \textbf{test} & \textbf{val} & \textbf{Visual Constr.} & \textbf{Multi-agent} & \textbf{Ambiguous} & \textbf{val} \\
\hline
AttnGrounder~\cite{mittal2020attngrounder}  & ResNet-50   & 61.32 & 62.34 & 64.35 & 62.31 & 64.57 & 62.74 & 64.82 & 64.31 & 57.25 \\
CMSVG~\cite{rufus2020cosine}        & EfficientNet& 68.61 & 67.66 & 68.47 & 68.87 & 69.93 & 69.39 & 66.77 & 67.83 & 62.21 \\
TransVG~\cite{deng2021transvg}      & ResNet-101  & 65.83 & 68.14 & 70.85 & 66.52 & 68.42 & 68.12 & 66.34 & 69.25 & 65.45 \\
CMRT~\cite{luo2020c4av}        & ResNet-152  & 69.11 & 69.42 & 68.83 & 69.54 & 70.37 & 67.12 & 66.20 & 62.23 & 64.25 \\
MDERT~\cite{kamath2021mdetr}        & ResNet-101  & 70.52 & 66.74 & 70.23 & 71.35 & 72.15 & 68.35 & 65.37 & 68.38 & 62.72 \\
VL-BERT~\cite{dai2020commands}      & ResNet-101  & 70.03 & 71.42 & 70.54 & 71.47 & 72.36 & \color{brown}\underline{70.29} & 70.14 & 69.84 & 66.70 \\
RSD-LXMERT~\cite{chan2022grounding}   & ResNet-101  & 72.64 & 72.35 & 71.46 & 73.37 & 74.52 & 70.22 & \color{brown}\underline{71.87} & 63.44 & 65.80 \\
VLTVG~\cite{yang2022improving}        & ResNet-101  & 63.33 & 67.14 & 68.26 & 65.37 & 68.49 & 68.51 & 66.22 & 70.24 & \color{brown}\underline{68.80} \\
Grounding-DINO~\cite{liu2023grounding} & ViT     & 68.15 & 67.92 & 68.48 & 69.50 & 70.10 & 66.17 & 65.85 & 67.24 & 63.15 \\ 
UNINEXT~\cite{yan2023universal}       & ResNet-50   & 70.87 & 70.62 & 71.34 & 71.35 & 73.47 & 69.26 & 68.78 & \color{brown}\underline{71.29} & 65.32 \\
CAVG~\cite{liao2024gpt}          & ViT         & \color{brown}\underline{74.62} & \color{brown}\underline{72.44} & \color{brown}\underline{73.25} & \color{brown}\underline{75.52} & \color{brown}\underline{76.48} & 68.39 & 67.36 & 69.45 & 64.36 \\
\hline
Qwen2.5-VL-7B~\cite{bai2025qwen2} & VLM     & 47.31 & 48.20 & 49.10 & 50.06 & 50.84 & 45.12 & 46.37 & 47.05 & 41.92 \\ 
Qwen2.5-VL-72B~\cite{bai2025qwen2} & VLM     & 56.17 & 57.10 & 57.85 & 58.92 & 59.74 & 53.43 & 54.25 & 55.17 & 49.83 \\ 
Qwen3-VL-8B~\cite{yang2025qwen3} & VLM     & 56.19 & 57.25 & 58.16 & 59.05 & 59.85 & 53.55 & 54.49 & 55.25 & 50.13 \\ 

\hline

\multirow{2}{*}{\textbf{E3AD (Ours)}} & \cellcolor{gray!10}VLM    & \cellcolor{gray!10} \textbf{80.12} & \cellcolor{gray!10}  \textbf{80.94} & \cellcolor{gray!10} \textbf{79.64} & \cellcolor{gray!10} \textbf{81.02} & \cellcolor{gray!10} \textbf{82.56} & \cellcolor{gray!10} \textbf{76.62} & \cellcolor{gray!10} \textbf{77.24} & \cellcolor{gray!10} \textbf{77.05} & \cellcolor{gray!10} \textbf{77.86} \\
 & \cellcolor{gray!25} -- & \cellcolor{gray!25} $\textbf{6.86\%}\uparrow$ & \cellcolor{gray!25} $\textbf{10.50\%}\uparrow$ & \cellcolor{gray!25} $\textbf{8.72\%}\uparrow$ & \cellcolor{gray!25} $\textbf{6.79\%}\uparrow$ & \cellcolor{gray!25} $\textbf{7.36\%}\uparrow$ & \cellcolor{gray!25} $\textbf{8.26\%}\uparrow$ & \cellcolor{gray!25} $\textbf{6.95\%}\uparrow$ & \cellcolor{gray!25} $\textbf{7.48\%}\uparrow$ & \cellcolor{gray!25} $\textbf{11.63\%}\uparrow$ \\
\toprule
\end{tabular}
}
\label{tab:vg}
\vspace{-6mm}
\end{table*}

\section{Experiments}
\noindent\textbf{Datasets.}
This study conducts experiments on several challenging real-world benchmarks, including Talk2Car~\cite{grujicic2022predicting}, DrivePilot~\cite{liao2025Think}, MoCAD~\cite{liao2024gpt}, and Talk2Car-Trajectory~\cite{deruyttere2022talk2car}. To further assess model robustness, we follow the ThinkDeeper protocol~\cite{liao2025Think} and introduce refined data splits for the DrivePilot and MoCAD, resulting in two tailored subsets: Long-Text and Corner-Case, each presenting distinct challenges.

\noindent\textbf{Evaluation Metrics.}
We evaluate models along two axes: (1) joint end-to-end performance and (2) sub-task ablations. For end-to-end evaluation, we report trajectory metrics, including ADE, FDE, Fréchet, DTW, SSPD, and planning accuracy within $g$ meters  (PA$_g$). For sub-task evaluation, we follow the Talk2Car C4AV protocol \cite{deruyttere2019talk2car} and report IoU for visual grounding; MAE and IoU for spatial reasoning; and Spearman’s ($\rho$) and Kendall’s ($\tau$) for emotion awareness.

\noindent\textbf{Implementation Details.} We train E3AD using the MS-Swift library~\cite{zhao2024swift} with LoRA fine-tuning~\cite{TL:LoRA} (rank 16, scaling factor 32), a constant learning rate of $10^{-4}$, and a per-device batch size of 16 for one epoch over the full dataset. Training is conducted in two stages: (i) modality pretraining and unified fine-tuning to adapt the backbone to domain data, and (ii) behavioral alignment via DPO to refine responses toward the desired driving behavior. All experiments are run on 8$\times$NVIDIA H200 GPUs. 
To ensure a fair comparison, we freeze the Qwen2.5-VL-7B backbone and train only low-rank adapters, keeping the trainable parameter budget comparable to or smaller than the baselines. The gains in Tables \ref{tab:baselines}-\ref{tab:vg} thus mainly stem from the proposed emotion modeling, dual-path spatial reasoning, and consistency-oriented training rather than model size. Larger generic VLMs (Qwen2.5-VL-72B, Qwen3-VL-8B) still underperform E3AD, indicating that task-aligned structure and objectives matter more than raw capacity. 

\subsection{Joint Evaluation Results}
\vspace{-2mm}
Table~\ref{tab:baselines} reports the end-to-end performance of E3AD and SOTA baselines under unified evaluation settings. In our formulation, emotion perception, spatial understanding, and visual grounding jointly provide intermediate evidence for downstream planning, simulating a more realistic AD pipeline that couples perception, cognition, and action. The results clearly show that E3AD outperforms all baselines across seven standard trajectory metrics. Compared with the strongest baseline (PTPC), our model achieves notable gains of 17.01\%, 18.26\%, and 20.00\% reductions in ADE, Fréchet, and FDE, respectively, indicating more accurate and stable trajectory forecasting. Likewise, E3AD improves SSPD and DTW by over 16\%, reflecting better temporal smoothness and spatial coherence in path generation. In planning accuracy, E3AD improves PA$_2$/PA$_4$ by +16.71\%/+18.10\%, indicating tighter alignment between plans and driving goals. General-purpose VLMs (Qwen-VL) still struggle, and FSDrive-Finetuned trails markedly, with higher FDE (10.45 vs. 6.64), confirming the limitations of existing VLA models in handling command-driven scenarios. Overall, E3AD’s stronger fusion of command semantics, spatial context, and emotion yields more reliable and command-aligned waypoint planning.

\subsection{Sub-task Evaluation Results}
We report the performance of E3AD on key sub-tasks against SOTA methods. Notably, our model addresses all tasks concurrently within one end-to-end trained network, whereas other baselines are specialized for only one task.

\noindent\textbf{Visual Grounding.} Table~\ref{tab:vg} reports visual grounding results. E3AD surpasses the strongest baseline (CAVG) with absolute gains of +6.86\% (Talk2Car), +10.50\% / +8.72\% (MoCAD test/val), and +6.79\% / +7.36\% (DrivePilot test/val). This superiority is especially pronounced in challenging scenes. On corner-case splits (occluded, multi-agent, ambiguous), it improves by +8.26\%, +6.95\%, and +7.48\%, and on Long-text by +11.63\%. Generic VLMs (Qwen) lag markedly across all datasets. These results indicate our emotion- and spatial-aware grounding yields more precise and robust localization within a unified E2E model.

\noindent\textbf{Spatial Reasoning.} Table~\ref{tab:Spatial} illustrates that E3AD markedly outperforms large VLM baselines (Qwen2.5/3-VL) on target localization and depth estimation. While even VLMs like Qwen2.5-VL-72B struggle with basic spatial reasoning (Location MAE 10.1, Depth MAE 22.68), E3AD demonstrates a precise, grounded understanding with an MAE of only 0.47 (Location) and 4.25 (Depth), respectively. This superiority is consistent in accuracy metrics, where E3AD achieves 97.7\% PA$_2$ (Location) and 53.1\% PA$_2$ (Depth), far above the best VLM. These results demonstrate that E3AD establishes a new state-of-the-art in 3D spatial perception and localization, significantly outperforming conventional visual-spatial VLMs in real-world driving scenarios.

\renewcommand{\arraystretch}{1.1}
\begin{table}[t]
\centering
\caption{Emotion prediction across valence, arousal, and dominance. Reported metrics are Spearman’s $\rho$ and Kendall’s $\tau$ correlations with ground-truth VAD ($\uparrow$: a higher value is better).}
\vspace{-3.3mm}
\resizebox{\linewidth}{!}{
\begin{tabular}{l|cc|cc|cc}
\bottomrule
\multirow{2}{*}{\textbf{Model}} 
& \multicolumn{2}{c}{\textbf{Valence $\uparrow$}} 
& \multicolumn{2}{c}{\textbf{Arousal $\uparrow$}} 
& \multicolumn{2}{c}{\textbf{Dominance $\uparrow$}} \\ \cmidrule(lr){2-3} \cmidrule(lr){4-5}  \cmidrule(lr){6-7}
& $\rho$ & $\tau$ & $\rho$ & $\tau$ & $\rho$ & $\tau$ \\ 
\hline
BERT~\cite{devlin2019bert} + Ridge    & 0.78 & 0.59 & 0.75 & 0.56 & 0.74 & 0.55 \\
RoBERTa~\cite{liu2019roberta} + Ridge   & 0.80 & 0.61 & 0.77 & 0.59 & 0.78 & 0.58 \\
DistilBERT~\cite{sanh2019distilbert} + Ridge   & 0.82 & 0.64 & 0.79 & 0.61  & 0.79  & 0.60 \\
Qwen2.5-7B-Instruct~\cite{bai2025qwen2}   & 0.11 & 0.08 & 0.02 & 0.02 & 0.04 & 0.03 \\
Qwen3-Emb.-4B~\cite{zhang2025qwen3} + Ridge  & 0.83  & 0.64  & 0.79  & 0.61  & 0.82 & 0.63 \\ 
\hline
\rowcolor{gray!15} \textbf{E3AD (ours)}  & \textbf{0.95} & \textbf{0.84} & \textbf{0.94} & \textbf{0.82} & \textbf{0.94} & \textbf{0.81} \\ \toprule
\end{tabular}
}
\label{tab:vad}
\vspace{-3.9mm}
\end{table}

\begin{table}[t]
\centering
\caption{Spatial reasoning results on Talk2Car vs. VLM baselines.}
\vspace{-3.2mm}
\resizebox{\linewidth}{!}{
\begin{tabular}{l|c|ccc|ccc}
\bottomrule
\multirow{2}{*}{\textbf{Model}}
               & \textbf{T2C} & \multicolumn{3}{c}{\textbf{Target Loc. Est.}}         & \multicolumn{3}{c}{\textbf{Target Depth Est.}}        \\ \cmidrule(lr){2-2}\cmidrule(lr){3-5}\cmidrule(lr){6-8}
               & IoU$_{50}$ & MAE $\downarrow$   & PA$_{2}$ & PA$_{4}$ & MAE $\downarrow$  & PA$_{2}$ & PA$_{4}$ \\ \hline
Qwen2.5-VL-7B~\cite{bai2025qwen2}  & 40.23    & 3.49  & 39.5          & 71.1          & 22.92 & 1.3           & 4.2           \\
Qwen2.5-VL-72B~\cite{bai2025qwen2} & 51.42    & 10.1  & 38.5          & 77.9          & 22.68 & 1.5           & 4.5           \\
Qwen3-VL-8B~\cite{yang2025qwen3}    & 52.68    & 3.71  & 32.8          & 71.5          & 18.89 & 13.7          & 26.5          \\ \hline
\rowcolor{gray!15} \textbf{E3AD (Ours)}        & \textbf{79.32}    & \textbf{0.47}  & \textbf{97.7}          & \textbf{98.8}          & \textbf{4.25}  & \textbf{53.1}          & \textbf{71.2}    \\ \toprule     
\end{tabular}
}
  \vspace{-6mm}
\label{tab:Spatial}
\end{table}

\noindent\textbf{Emotion Recognition.}
Table~\ref{tab:vad} presents emotion recognition results across valence, arousal, and dominance dimensions. 
Since VAD represents relative emotional magnitudes, we adopt Spearman's rank correlation ($\rho$) and Kendall’s rank correlation coefficient ($\tau$) as evaluation metrics to assess the monotonic relationship between model predictions and labels. 
The \textit{Ridge} regressor obtains command embeddings from language models and applies ridge regression to predict continuous VAD values, serving as a lightweight emotion estimator. E3AD achieves the highest correlation with human annotations, reaching 0.95/0.84 ($\rho$, $\tau$) for valence, 0.94/0.82 for arousal, and 0.94/0.81 for dominance. In contrast, Qwen2.5-7B and Qwen3-4B  exhibit near-random correlation levels, revealing limited sensitivity to emotion cues. These results show that E3AD effectively models the continuous emotion space of human emotions, capturing valence (positive-negative sentiment), arousal (intensity), and dominance (sense of control), enabling human-centric planning and verbal feedback for passengers.

\subsection{Ablation Study} 
Table~\ref{Ablation_1} summarizes module contributions. Removing the egocentric pathway causes the largest VG drop ($\downarrow$7.0\% IoU on Talk2Car; $\downarrow$6.6\% on Vision-Constraint), confirming its role in first-person grounding that aligns linguistic references with observed targets. Excluding the allocentric map weakens global reasoning by removing complementary spatial semantics and scene topology. The Emotion Modeling shows the greatest benefit on ambiguous and long-text commands ($\uparrow$4.5\%/$\uparrow$4.8\%), enhancing the model’s sensitivity to nuanced linguistic cues and emotionally rich expressions. DPO gives moderate gains by refining multimodal alignment.
For waypoint planning (Table~\ref{Ablation_2}), the allocentric map is critical: removal degrades trajectories (ADE/FDE ($\uparrow$10.0\%/$\uparrow$10.1\%)), indicating the value of global priors for spatial awareness and route consistency. The egocentric pathway anchors local motion cues; Emotion Modeling and DPO further improve emotion consistency.

\renewcommand{\arraystretch}{1.085}
\begin{table}[t]
  \centering
  \caption{Ablation study of E3AD's core components on visual grounding performance on the Talk2Car (T2C) benchmark, vision constraint (Constr.), ambiguous (Ambg.), and long-context command (Long) test sets. Components: Egocentric pathway (Ego.), Allocentric pathway (Allo.), DPO, and Emotion Modeling (Emo.).}
  \vspace{-3mm}
      \resizebox{\linewidth}{!}{
    \begin{tabular}{cccc|cccc}
     \hline
      
    \textbf{Ego.}  & \textbf{Allo.} & \textbf{Emo.}  & \textbf{DPO}   & \textbf{T2C} $\uparrow$   & \textbf{Constr.} $\uparrow$  & \textbf{Ambg.} $\uparrow$ & \textbf{Long} $\uparrow$ \\
     \hline
    \xmark     & \cmark     & \cmark     & \cmark     & 74.48  & 71.60    & 72.24  & 72.47  \\
    \cmark     & \xmark     & \cmark     & \cmark     & 76.48  & 73.92    & 74.65  & 74.76  \\
    \cmark     & \cmark     & \xmark     & \cmark     & 78.78  & 74.41    & 73.57  & 74.12  \\
    \cmark     & \cmark     & \cmark     & \xmark     & 79.55  & 75.58    & \textbf{77.09} & 76.44  \\
    \hline
   \rowcolor[gray]{0.9} \cmark     & \cmark     & \cmark     & \cmark     & \textbf{80.12} & \textbf{76.62}  & 77.05  & \textbf{77.86} \\
        \hline
    \end{tabular}%
    }
    \label{Ablation_1}%
  \vspace{-3.5mm}
\end{table}%

\renewcommand{\arraystretch}{1.05}
\begin{table}[t]
  \centering
\caption{Ablation of core designs in E3AD on waypoint planning.}
  \vspace{-3mm}
      \resizebox{\linewidth}{!}{
    \begin{tabular}{cccc|cccc}
   \hline
        
    \textbf{Ego.}  & \textbf{Allo.} & \textbf{Emo.}  & \textbf{DPO}   & \textbf{ADE} $\downarrow$   & \textbf{SSPD} $\downarrow$  & \textbf{Fréchet} $\downarrow$     & \textbf{FDE} $\downarrow$ \\
    \hline
    \xmark     & \cmark     & \cmark     & \cmark     & 4.12  & 2.06  & 7.61  & 7.02  \\
    \cmark     & \xmark     & \cmark     & \cmark     & 4.27  & 2.15  & 7.86   & 7.31  \\
    \cmark     & \cmark     & \xmark     & \cmark     & 3.93  & 1.91  & 7.36   & 6.80  \\
    \cmark     & \cmark     & \cmark     & \xmark     & 3.96  & 1.89  & 7.31   & 6.86  \\
    \hline
   \rowcolor[gray]{0.9}  \cmark     & \cmark     & \cmark     & \cmark    & \textbf{3.88} & \textbf{1.86} & \textbf{7.23}& \textbf{6.64} \\
     \hline
    \end{tabular}%
    }%
  \label{Ablation_2}
  \vspace{-2mm}
\end{table}%

\begin{figure}[t]
    \centering
    \includegraphics[width=.9\linewidth]{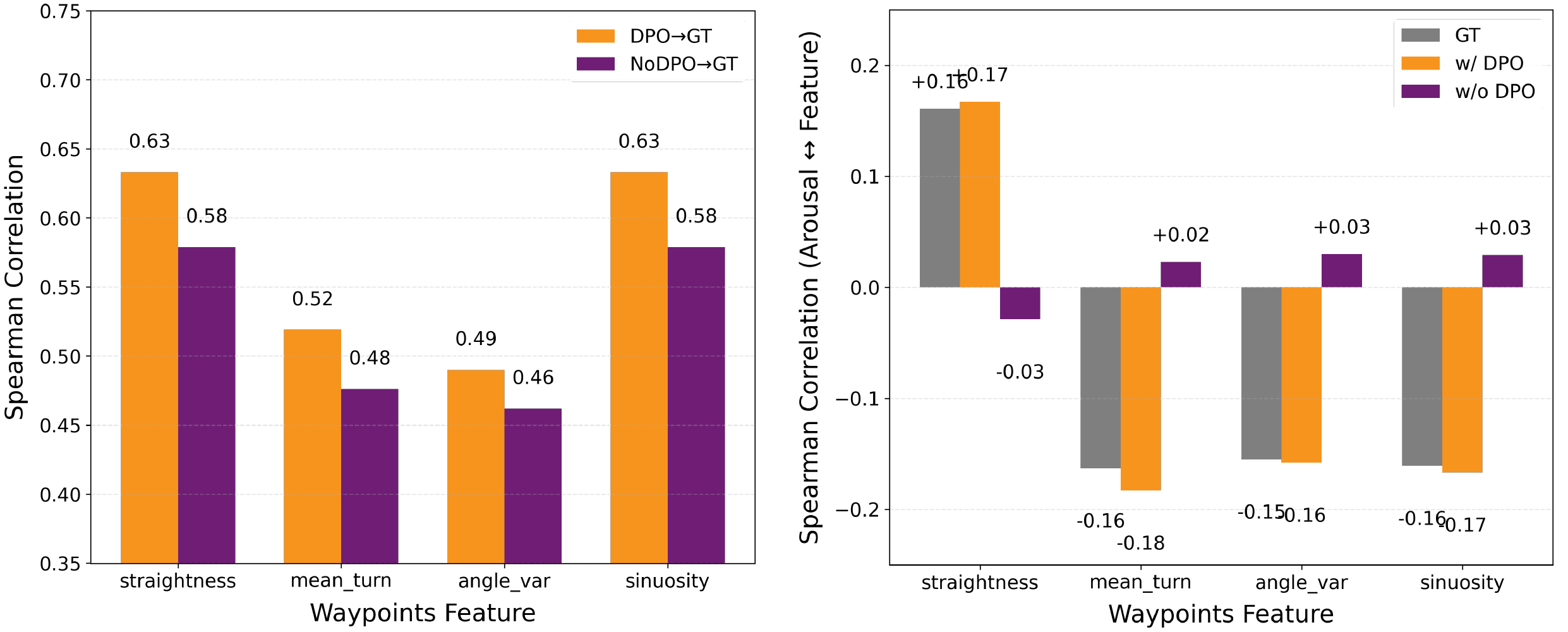}
    \vspace{-3mm}
    \caption{DPO’s effect on emotion-trajectory consistency.}
    \label{fig:dpo}
    \vspace{-7mm}
\end{figure}

\begin{figure*}[htb]
    \centering
    \includegraphics[width=\linewidth]{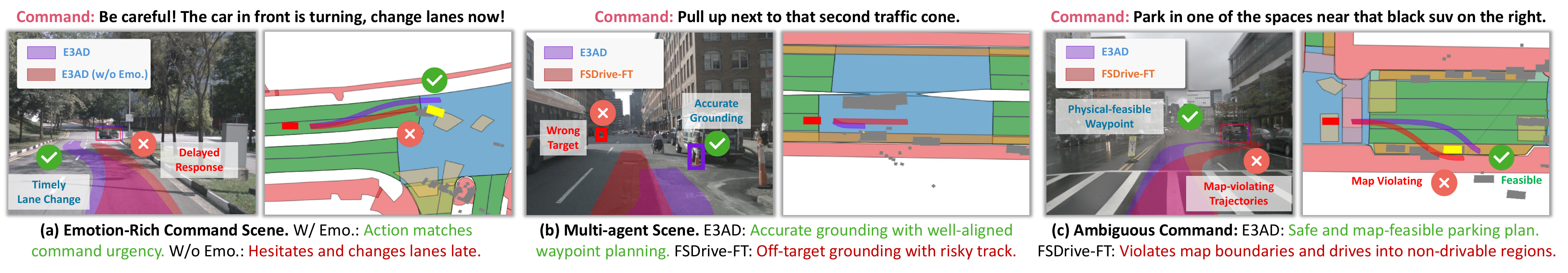}
    \vspace{-8mm}
    \caption{Qualitative comparison between E3AD and FSDrive-FT in emotion-rich (a), multi-agent (b), and ambiguous (c) scenes.}
    \label{fig:qualitative}
\vspace{-5mm}
\end{figure*}

As in Fig.~\ref{fig:dpo}, DPO raises Spearman correlations for straightness/sinuosity and stabilizes turn smoothness and angular variation, yielding geometrically coherent, behaviorally consistent paths. Higher arousal corresponds to straighter, smoother motion; lower arousal to more cautious, curved motion. Overall, DPO strengthens emotion-trajectory consistency even when numeric gains are modest.

\subsection{Case Study} 
\label{sec:casestudy}

\begin{figure}[t]
    \centering
    \vspace{-0.5mm}
    \includegraphics[width=\linewidth]{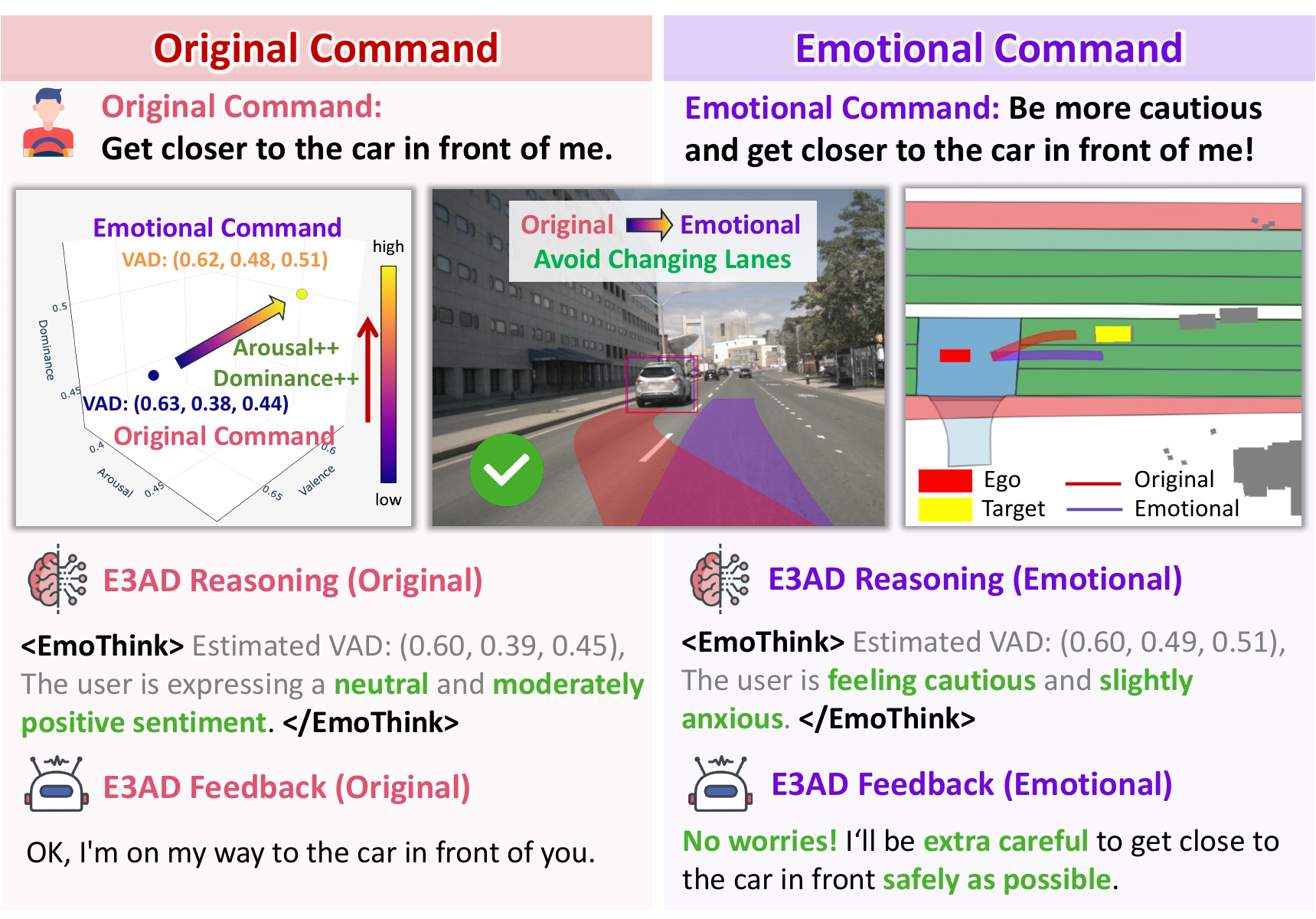}
    \vspace{-5mm}
    \caption{Case study for the impact of different emotional commands (neutral vs. cautious) on E3AD’s end-to-end performance.}
    \label{case_study}
\vspace{-8mm}
\end{figure}

Fig.~\ref{case_study} illustrates how E3AD integrates emotion understanding with spatial reasoning to produce end-to-end behavior. The VAD plot showcases that the Emotion Modeling component captures the linguistic shift from a neutral command to its cautious variant: the neutral command maps to $(0.60, 0.39, 0.45)$, while adding the qualifier ``Be more cautious'' shifts the VAD state to $(0.60, 0.49, 0.51)$, increasing Arousal and Dominance. Conditioned on the neutral VAD, E3AD plans a standard lane-change maneuver; conditioned on the cautious VAD, the DPO-aligned policy instead avoids the lane change altogether. The generated textual feedback is also conditioned on emotion state $\hat{e}$: the EmoThink block infers a cautious, slightly anxious passenger and produces a reassuring explanation. 
Taken together with the geometric statistics in Fig.~\ref{fig:dpo}, these results indicate that emotion supervision affects not only language but also motion geometry. At fixed intent, higher arousal steers E3AD toward straighter paths with less lateral oscillation and earlier hazard avoidance, while lower arousal produces slower approaches with larger safety margins. Thus, the VAD vector serves as a continuous control signal that selects among behaviorally distinct yet physically valid plans, rather than as a binary switch or style token.

\subsection{Qualitative Analysis} 
\label{sec:qualitative}
Fig.~\ref{fig:qualitative} showcases qualitative comparisons between E3AD and FSDrive-FT~\cite{zeng2025futuresightdrive} across three challenging scenes.
In the emotion-rich command scene (Panel (a)), E3AD leverages its continuous VAD-based emotion modeling to infer the passenger’s heightened urgency and produces a timely lane-change response. In contrast, the variant without emotion modeling hesitates and reacts too late. In the multi-agent scene (Panel (b)), where objects are partially occluded and visual cues are ambiguous, E3AD integrates allocentric topology with egocentric evidence through spatial reasoning, enabling accurate target grounding and a safe, well-aligned trajectory. FSDrive-FT fails to parse the complex scene structure, resulting in off-target grounding and unsafe plans. In the ambiguous command scene (panel (c)), E3AD resolves linguistic ambiguity by combining language tone with spatial context, producing a feasible, map-consistent parking path. In contrast, FSDrive-FT crosses map boundaries into non-drivable regions. Overall, this illustrates how emotion-aware language understanding and dual-frame spatial reasoning enable E3AD to deliver more interpretable and human-aligned grounding and planning. See Appendix~\textcolor{cvprblue}{C} for more qualitative results of E3AD.

\begin{figure}[h]
    \centering
    \vspace{-1mm}
    \includegraphics[width=\linewidth]{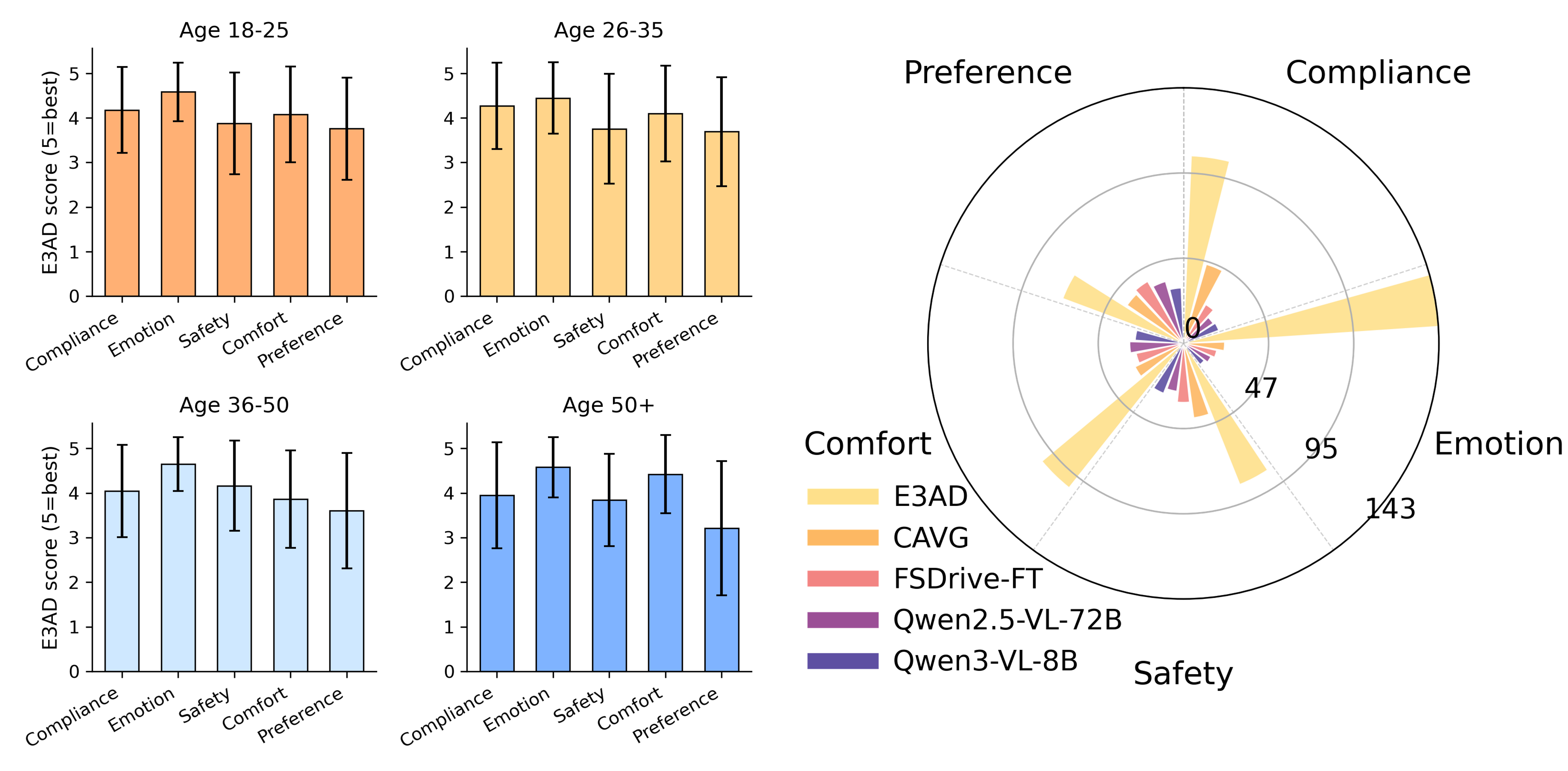}
    \vspace{-8.6mm}
    \caption{User study on perceived compliance, emotion, safety,  and preference. (Left) E3AD consistently achieves high Likert scores across all age groups. (Right) Comparison of Rank-1 votes, E3AD dominates in most dimensions, outperforming all baselines.}
    \label{fig:userstudymain}
    \vspace{-5mm}
\end{figure}
\subsection{User Study}

To evaluate the real-world utility of E3AD, we conduct a user study with 217 participants. Participants viewed anonymized clips from five systems and ranked them on Command Compliance, Emotion Alignment, Safety, Comfort, and Overall Preference. As shown in Fig.~\ref{fig:userstudymain}, E3AD obtains the highest mean scores and most Rank-1 votes on most dimensions, indicating a clear preference over the baselines. Beyond absolute scores, participants consistently prefer trajectories that match the command’s emotional tone, maintain correct grounding, and stay physically feasible. Systems without emotion modeling (FSDrive-FT, generic VLMs) are often seen as hesitant or unsafe, whereas E3AD’s emotion-aware grounding and map-feasible planning are judged more confident and trustworthy, highlighting the value of emotion cues in the end-to-end AD. 

\vspace{-1mm}
\section{Conclusion and Future Work}
\vspace{-1mm}
This work extends E2E AD from purely rational control to emotion-aware grounding and planning. We introduce OD-E2E AD and propose E3AD, a VLA framework that couples continuous VAD-based emotion modeling, dual-pathway spatial reasoning, and consistency-oriented training. Experiments on four real-world benchmarks show consistent improvements in visual grounding, trajectory quality, and continuous emotion estimation over strong baselines. In future work, we plan to incorporate more human preference feedback, richer multimodal emotion signals beyond language, and closed-loop evaluation in high-fidelity simulators to further enhance the reliability and acceptance of emotion-aware autonomous driving systems.

\section*{Acknowledgements}

This work was supported in part by the Natural Sciences and Engineering Research Council (NSERC) of Canada, and part by the Research Services and Knowledge Transfer Office, University of Macau [SRG2023-00037-IOTSC, MYRG-GRG2024-00284-IOTSC], in part by the State Key Lab of Intelligent Transportation System [2024-B001]. 


{
    \small
    \bibliographystyle{ieeenat_fullname}
    \bibliography{main}
}

\end{document}